\pdfoutput=1
\documentclass{article}

\usepackage[final,nonatbib]{neurips_2024}

\usepackage[utf8]{inputenc} 
\usepackage[T1]{fontenc}    
\usepackage{hyperref}       
\usepackage{url}            
\usepackage{booktabs}       
\usepackage{amsfonts}       
\usepackage{nicefrac}       
\usepackage{microtype}      
\usepackage{xcolor}         

\usepackage{amsthm}
\usepackage{tabularx}
\usepackage{multirow}
\usepackage{graphicx}
\usepackage[acronym,nomain]{glossaries}
\usepackage{tikz-cd}

\newacronym{sgd}{SGD}{Stochastic Gradient Descent}
\newacronym{gs}{GS}{Gumbel-Softmax}
\newacronym{cnn}{CNN}{Convolutional Neural Network}
\newacronym{dnn}{DNN}{Deep Neural Network}
\newacronym{relu}{ReLU}{Rectified Linear Unit}
\newacronym{flop}{FLOP}{Floating Point Operation}
\newacronym{cv}{CV}{Computer Vision}
\newacronym{cpu}{CPU}{Central Processing Unit}
\newacronym{gpu}{GPU}{Graphics Processing Unit}

\glsdisablehyper

\DeclareMathOperator*{\argmax}{arg\,max}

\newcolumntype{Y}{>{\centering\arraybackslash}X}

\newtheorem{theorem}{Theorem}[section]
\newtheorem{lemma}[theorem]{Lemma}

\title{Inducing Semi-Structured Sparsity by Masking for Efficient Model Inference in Convolutional Networks}

\author{%
David A. Danhofer
\\
Department of Computer Science\\
ETH Zürich\\
\texttt{ddanhofer@ethz.ch}\\
}

\begin{document}

\maketitle

\begin{abstract}
	The crucial role of convolutional models, both as standalone vision models and backbones in foundation models, necessitates effective acceleration techniques. This paper proposes a novel method to learn semi-structured sparsity patterns for convolution kernels in the form of maskings enabling the utilization of readily available hardware accelerations. The approach accelerates convolutional models more than two-fold during inference without decreasing model performance. At the same time, the original model weights and structure remain unchanged keeping the model thus easily updatable. Beyond the immediate practical use, the effect of maskings on prediction is easily quantifiable. Therefore, guarantees on model predictions under maskings are derived showing stability bounds for learned maskings even after updating the original underlying model.\footnote{Code available at \href{https://github.com/ddanhofer/Semi-Structured-Sparsity-CNNs}{github.com/ddanhofer/Semi-Structured-Sparsity-CNNs}
	}
\end{abstract}

\section{Introduction}\label{sec:introduction}

The increasing complexity of deep learning models \cite{justusPredictingComputationalCost2018},  their deployment in applications \cite{chaiDeepLearningComputer2021}, and the adoption of reflection incurring several inference passes per query, e.g., as in the O1 models from the GPT family \cite{brownLanguageModelsAre2020}, shifts the relative amounts of resources spent during the model lifetime from the training to the inference stage \cite{desislavovComputeEnergyConsumption2023, parkDeepLearningInference2018}. It therefore becomes imperative to make models more efficient \cite{zhangAdvancingModelPruning2022}. One way of achieving this is by spending a comparatively small, additional share of resources during training to learn a one-time modification of the model that lowers the model's inference and thus lifetime cost \cite{menghaniEfficientDeepLearning2023, thompsonComputationalLimitsDeep2023a}. First and foremost, such a modification is effective if it decreases the model's computational and time cost at a relatively low additional training overhead while not affecting the prediction performance of the model negatively \cite{liEvaluatingEnergyEfficiency2016}. Additionally, there are other desirable properties of such one-time modifications: From an application perspective the achievable gain in efficiency is only useful if it can be leveraged easily, a well-known challenge, e.g., with sparsifying models \cite{galeStateSparsityDeep2019, hookerHardwareLottery2021}. Taking into consideration the increasing popularity of large, expensive to train, foundation models \cite{huLoRALowRankAdaptation2021} or models employed in an online setting subject to continuous updates the proposed change should not affect the possibility to update the model, e.g., by changing the weights or architecture underlying the model. Ideally, if such a model is updated,  the learned modification can even be reused under the constraint of the magnitude of change imposed by updating the model.

\textit{Semi-structured sparse maskings} satisfy the above properties by replacing the dense matrix operations usually required during inference by cheaper and faster operations on semi-structured sparse matrices \cite{bulucParallelSparseMatrixMatrix2012}. While many works have demonstrated that sparse (pruned) submodels can solve the same task at almost no loss of performance \cite{blalockWhatStateNeural2020a, liuRethinkingValueNetwork2019} the sparsity of the models does not necessarily have to adhere to a specific pattern making it difficult to turn theoretically obtained computational speedups by saving on data loading and computational operations into practical efficiency gains \cite{hoeflerSparsityDeepLearning2021}. Regular patterns are more ``machine-friendly'' inducing the desired efficiency \textit{a priori} but limiting the choices for the sparse patterns, which thus need to be chosen carefully with the goal of minimizing the loss of inference performance in mind.

This paper proposes a novel method of learning regularly sparse masking patterns for convolutions, key building blocks for state-of-the art \gls{cv} models \cite{liuConvNet2020s2022} and foundation models building on \gls{cv} models as their backbone \cite{radfordLearningTransferableVisual2021a}. The proposed method
\begin{itemize}
	\item shows how to effectively use readily available hardware accelerations for semi-structured sparse matrices in convolution kernels to accelerate inference,
	\item outperforms available heuristics for semi-structured sparsity showing that semi-structured sparsity masks can be learned with a fraction of the original training resources while incurring a negligible performance loss in \gls{cv} classification tasks,
	\item provides the additional advantage of not changing the original set of trained weights keeping models updatable and rendering the method especially attractive for use in large models, e.g., foundation models and in online settings,
	\item induces an easily quantifiable change to the model's prediction behavior and thus lends itself to settings where hard guarantees on model predictions are of interest.
\end{itemize}

In the following section the adoption of semi-structured sparsity and sparsity in convolutional models are addressed. Section \ref{sec:methods} of the paper covers modeling semi-structured sparsity in general, in convolutional models, and the theoretical implications of such model alterations in inference. The results of empirically testing the method on widely used convolutional architectures are presented in Section \ref{sec:results} followed up by a discussion of the method presented and a conclusion.

\section{Related Work}\label{sec:background}

In the following the notion and adoption of semi-structured sparsity is introduced. Then, the implications on prediction performance and the computational challenges of sparsifying \glspl{cnn} are highlighted.

\begin{figure}[ht]
	\centering
	\def\svgwidth{\linewidth}
	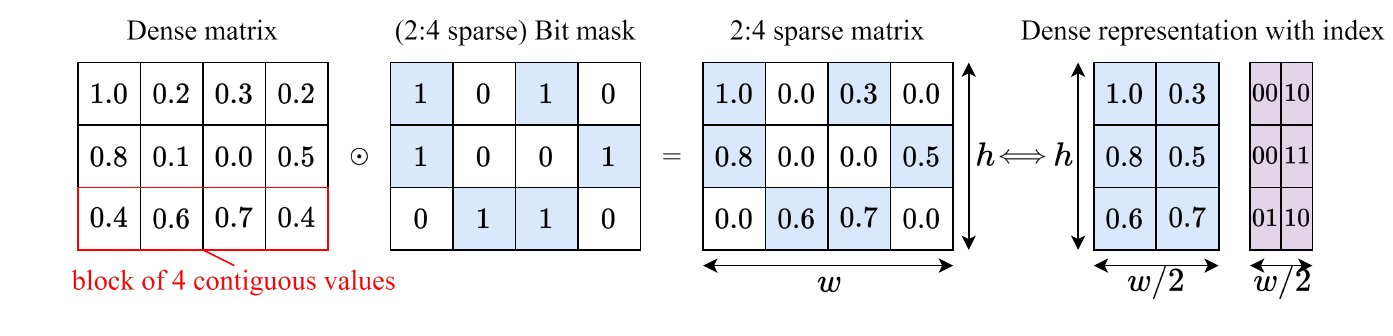
	\caption{A 2:4 sparse matrix of floating point values -- obtained from a dense matrix and a sparse bit mask -- and its equivalent structured representation containing only the non-zero entries and a 2-bit index preserving the structure taking up roughly only half the space.}
	\label{fig:24matrix}
\end{figure}

\paragraph{Semi-Structured Sparsity} Semi-structured sparsity to accelerate network inference has been introduced in \cite{poolChannelPermutationsSparsity2021} as \textit{N:M-sparsity} requiring \(N\) out of \(M\) elements of a contiguous block to be zero (s.\ a. \ref{subsec:sparsity}). Beyond the general case of N:M sparsity, the practically interesting special case of 2:4 sparsity has been considered in more detail \cite{huangPruningLargeLanguage2024, hubaraAcceleratedSparseNeural2021} in which exactly half of the weights are pruned as illustrated in Figure \ref{fig:24matrix}. This setting enables hardware acceleration via NVIDIA sparse tensor cores available from the NVIDIA Ampere architecture on via the TensorRT v8.0 library \cite{poolAcceleratingInferenceSparsity2021}. Since half the elements are zeroed out and thus negligible, the amount of data to load from memory is almost halved with the number of \glspl{flop} needed to conduct an operation on the sparse matrix also decreasing, e.g., linearly for addition and element-wise operations and super-linearly for multiplication, decomposition etc. \cite{yusterFastSparseMatrix2005}. This way 2:4 sparse matrix operations compute the same effective operation while reducing the time needed by a factor of two \cite{poolAcceleratingInferenceSparsity2021}. The difficulty in turning a dense matrix into a 2:4 sparse matrix, however, lies in selecting the most useful two of the four weights in each quadruple. To this end \cite{poolChannelPermutationsSparsity2021} propose a permutation regime that allows for preserving the weights based on magnitude and assess the found pattern via a scoring mechanism, the efficacy score. The functionality is available via NVIDIA's Apex library \cite{nvidiacorporationApexPyTorchExtension2018}. Notably, pruning via Apex requires finetuning the network again after pruning to achieve an inference performance comparable to that of the dense network in \gls{cv} tasks, e.g., classification \cite{poolAcceleratingInferenceSparsity2021, poolChannelPermutationsSparsity2021}, and therefore changes the original pretrained weights.

\begin{figure}[ht]
	\centering
	\def\svgwidth{0.95\linewidth}
	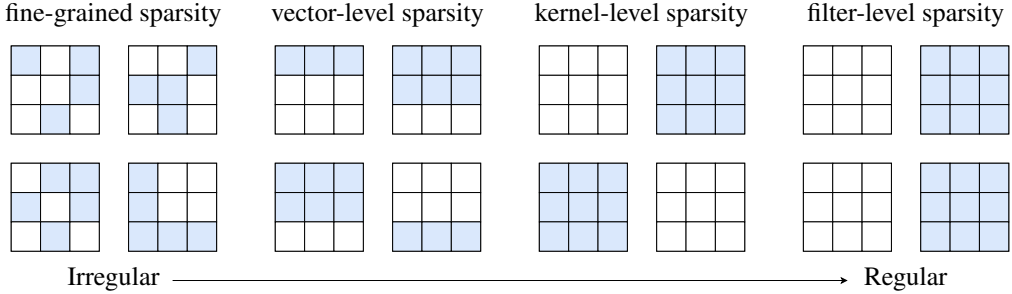
	\caption{Different levels of granularity in a 4D-tensor as used in 2D-convolutions of multi-channel inputs with several filters; although the same number of weights is retained and pruned across all levels of structure the ease of processing increases as structure increases but limits the number of possible patterns at the same time.}
	\label{fig:levels_of_sparsity}
\end{figure}

\paragraph{Sparsity in \glspl{cnn}}
The state-of-the-art performance of \glspl{cnn} in image-based and other tasks comes at the cost of a large memory footprint and computational cost. Pruning to obtain sparse networks is therefore a popular technique to decrease the computational and storage cost of deep neural networks \cite{blalockWhatStateNeural2020a}. Pruning techniques include magnitude-based pruning \cite{hanLearningBothWeights2015}, iterative pruning \cite{youDrawingEarlyBirdTickets2022}, and dynamic pruning \cite{linDynamicModelPruning2020}. Although theoretically any sparsity reduces the computational costs of such networks, irregularity in the sparsity patterns makes it difficult to map the required computations to (parallel) processor operations \cite{iofinovaHowWellSparse2022}. Even extremely high levels of (irregular) sparsity, i.e., \(> 97\%\), often yield no inference acceleration suffering from lack of library support \cite{galeStateSparsityDeep2019, wenLearningStructuredSparsity2016}. As visualized in Figure \ref{fig:levels_of_sparsity}, in the case of \glspl{cnn} different granularities of sparsity emerge naturally with more regular patterns being more ``machine-friendly'' effectively inducing smaller, still dense models \cite{hoeflerSparsityDeepLearning2021}. Structured pruning approaches pruning filters or even entire channels at once \cite{liPruningFiltersEfficient2017, molchanovPruningConvolutionalNeural2017, novaGradientfreeStructuredPruning2023}, however, quickly deteriorate prediction performance \cite{grimaldiAcceleratingDeepNeural2023, maoExploringGranularitySparsity2017, maoExploringRegularitySparse2017}. This motivates semi-structured sparsity, a fine-grained yet structured sparsity pattern, to maintain a large degree of freedom in the selection of sparsity patterns to not impede performance while also observing some (machine-)usable regular structure.

\section{Methods}\label{sec:methods}
In the following, the concept of modeling semi-structured sparsity in a network architecture and its effects on classification are introduced. Then, its application to convolutions is detailed out.

\subsection{Modeling Semi-Structured Sparsity}\label{subsec:sparsity}
N:M sparsity divides a matrix into non-overlapping blocks of \(M\) contiguous elements requiring that \(N\) of these elements be zero as these elements can subsequently be ignored in many matrix operations, e.g., addition or multiplication. 
There are exactly \(n = {(M-N)!}/{N!}\) ways of choosing \(N\) of the \(M\) elements without replacement and ignoring permutations, yielding \(n\) unique sparsity patterns. Selecting one of the \(n\) patterns can be modeled via a categorical variable \(z\) with class probabilities \(\pi_1, ..., \pi_n\) s.t. each probability denotes the probability of selecting the corresponding N:M sparsity pattern. Sampling the choice vector \(z\), a \(n\)-dimensional one-hot vector on the simplex \(\Delta^{n-1}\), from such a categorical distribution can be performed efficiently via the Gumbel-Max trick \cite{gumbelStatisticalTheoryExtreme1954}
\begin{equation}\label{equ:discr_z}
	z = \text{onehot}(\argmax_{i} [g_i + \log \pi_i])
\end{equation}
where \(g_i \sim \text{Gumbel}(0,1)\).
Aggregating all \(n\) N:M sparsity patterns as column vectors in a pattern matrix \(D \in \{0,1\}^{N\times n}\) allows for constructing the (row major) semi-structured sparse mask \(M\) of dimensions \(h\times Nw\) with \(h,w\in \mathbb{N}\) from one-dimensional block entries sampled from independent categorical distributions.
\begin{equation}\label{equ:sample_m}
	M = \begin{pmatrix}
		b^T & \dots\\
		\vdots & \ddots
	\end{pmatrix} \in \{0,1\}^{h\times Nw}\quad \text{where}\quad b = Dz \in \{0,1\}^{N},\ z \sim \text{Categorical}(\pi_1, ..., \pi_n)
\end{equation}
Since the \(\argmax\) operator employed to sample \(z\) according to (\ref{equ:discr_z}) is discrete and thus not differentiable this method cannot be used directly in a neural network to optimize the parameters \((\pi_i)_{i\in [n]}\) via backpropagation. Instead a differentiable approximation is constructed by replacing the \(\argmax\) operator with a softmax. The choice vector \(z\) can now be drawn as follows
\begin{equation}\label{equ:soft_gs}
	z_i = \frac{\exp((g_i + \log \pi_i) / \tau)}{\sum_k \exp((g_k + \log \pi_k) / \tau)}
\end{equation} 
yielding a \gls{gs} distribution \cite{jangCategoricalReparameterizationGumbelSoftmax2017} over the \(n\) choices additionally parameterized by the temperature parameter \(\tau\).
While not identical to a categorical distribution, the \gls{gs} distribution approximates a categorical distribution over the choices for small temperature values, e.g., \(\tau = 0.1\). This distribution allows for expressing the gradient as a deterministic function of the choice weights \(\pi\) and an independent source of random noise and thus gradient propagation through the stochastic node \cite{jangCategoricalReparameterizationGumbelSoftmax2017}. By updating the class probabilities in the distribution in respect to the classification objective the choice weights can be optimized for the classification task and the optimal choice is selected. After convergence the choice weights are frozen and the inference bit mask is obtained by sampling the blocks one final time from the \gls{gs} distributions yielding the sparse bit mask. An element-wise multiplication of the bit mask with the dense weight matrix results in the desired semi-structured sparse matrix and, by extension, \gls{cnn}.

\subsection{Effects of Maskings on Classifier Class Predictions}\label{subsec:effects}
Understanding how sparsity-inducing maskings affect a classifier's predictions is crucial to effectively trade off inference acceleration and potential performance inhibitions. The following Lemma contains the classifier definition  and states a useful property. This definition is used in all subsequent theoretical results and models the architectures considered for experiments closely. All proofs in this section are deferred to Appendix \ref{appendix}.
\begin{lemma}\label{lemma:classifier}
	Let \(f(x) = (\text{softmax} \circ f_d \circ ... \circ f_1)(x)\) be a compositional classifier of depth \(d\) with \(f_i(x) = \sigma(W_ix + b_i)\) predicting the class probabilities of an input sample \(x \in X\) across \(c\) classes. Let \(\sigma\) be a non-linear element-wise activation function and \(L\)-Lipschitz. Then \(f(x)\) is \(L_f\)-Lipschitz with \(L_f \leq L^d\prod_i\|W_i\|\). Let such a classifier be a compositional (\(L_f\)-)Lipschitz classifier.
\end{lemma}
Let the labels be one-hot elementary vectors \(e_i \in \mathbb R^c\) and let \(\lambda(\tilde y) = \min_i \|\tilde y - e_i\|_\infty\) be the function discretizing the classifier's probability prediction \(\tilde y \in \mathbb{R}^c, \sum_i \tilde y_i = 1, 0\leq \tilde y_i\leq 1\ \forall i\) into a class prediction. Given a prediction \(\tilde y \in \mathbb R^c\) by a classifier \(f(x)\) on a sample \(x \in X\) let the \textit{confidence} \(0 \leq \gamma_f(x) \leq \frac{1}{2} + \epsilon, \epsilon > 0\) of a classifier be defined as the minimum change \(\min_{\delta \in \mathbb R^c}\|\delta\|_\infty\) s.t. \(\tilde y + \delta\) is a valid probability vector and \(\lambda(\tilde y) \neq \lambda(\tilde y + \delta)\). Intuitively, this minimum change vector shifts the probability mass from the class with the highest probability to the class with the second highest probability to change the discretized class prediction \(\lambda(\cdot)\) with a minimal shift.
Let \(W\) be any weight matrix in a classifier and \(\Delta W\) an additive perturbation of the weights yielding \(W' = W + \Delta W\). Let a classifier be \textit{stable} in respect to a perturbation and a given sample \(x\) if the perturbation doesn't affect the classifiers prediction on \(x\), i.e., \(\lambda(f_W(x)) = \lambda(f_{W'}(x))\). 
\begin{lemma}\label{lemma:addit_perturb}
	Let \(W\) be any weight matrix in a compositional (\(L_f\)-)Lipschitz classifier \(f(x)\) and \(\Delta W\) an additive perturbation of the weights yielding \(W'_j = W_j + \Delta W\) and a perturbed compositional classifier \(f'(x)\). Then \(f'(x)\) is \(L_{f'}\)-Lipschitz with \(L_{f'} \leq L_f + L\ \|\Delta W \|\prod_{i=1, i\neq j}^{d}\|W_i\|\).
\end{lemma}
A masking of a matrix \(W\), i.e., zeroing out some (or all) of the entries, can be modeled as an element-wise product of the matrix \(W\) with a bit mask \(B\). However, any masking can always equivalently be described as an additive perturbation. Let \(\mu(B, W)\) be the masking function yielding this additive perturbation.
\begin{lemma}\label{lemma:norm_of_bitmask}
	For any bit mask \(B\) and a matrix \(W\) the additive perturbation \(\Delta W = \mu(B, W)\) s.t. \(W + \Delta W = B \odot W\) always exists and fulfills \(\|\Delta W\|_\infty \leq \|W\|_\infty\). The bound is tight.
\end{lemma}
The model of a compositional classifier from Lemma \ref{lemma:classifier} yields the following result guaranteeing stability of the classifier as a function of the perturbation and prediction confidence.
\begin{lemma}\label{lemma:additive_stability}
	Let \(W_j\) be the \(j\)-th weight matrix in a compositional Lipschitz classifier \(f(x)\) and \(\Delta W_j\) an additive perturbation of the weights \(W_j' = W_j + \Delta W_j\). Then the classifier is guaranteed to be stable in respect to such a perturbation and a sample \(x\) predicted with a confidence of \(\gamma_f(x) > L^{d}\ \|\Delta W\|\ \|x\|\prod_{i=1, i\neq j}^{d}\|W_i\|\).
\end{lemma}
Combining the statements made in Lemma \ref{lemma:additive_stability} above with the reformulation of maskings of weights as additive perturbations from Lemma \ref{lemma:norm_of_bitmask} yields the following result left without proof.
\begin{lemma}\label{lemma:masking_stability}
	Let \(W_j\) be the \(j\)-th weight matrix in a compositional Lipschitz classifier \(f(x)\) and \(\Delta W_j = \mu(B, W)\) an additive perturbation of the weights \(W_j' = W_j + \Delta W_j\) induced by any\ masking \(B\). Then the classifier is guaranteed to be stable in respect to such a perturbation and a sample \(x\) predicted with a confidence of \(\gamma_f(x) > L^{d}\ \|x\|\prod_{i=1}^{d}\|W_i\|\)
\end{lemma}
The looser bound in Lemma \ref{lemma:masking_stability} addresses the general case in which any masking is considered and thus a worst case assumption. For a specific masking or a constrained class of maskings a tighter bound as in Lemma \ref{lemma:additive_stability} can be derived. Lastly, consider the case in which one such specific masking has been learned for a classifier, which has since been updated, e.g., due to new data available. Applying the same masking to the updated classifier yields the following bound.
\begin{lemma}\label{lemma:combined_effect}
	Let \(W_j\) be the \(j\)-th weight matrix in a compositional Lipschitz classifier \(f(x)\), \(\Delta W_j = \mu(B, W)\) an additive perturbation of the weights \(W_j' = W_j + \Delta W_j\) induced by any masking \(B\) yielding the masked classifier \(f'(x)\). Let the update \(U_j\) be an additive perturbation of the weights \(V_j = W_j + U_j\) yielding the updated classifier \(f_U(x)\). Then the masked and updated classifier \(f_U'(x)\) obtained from applying the mask \(B\) to the updated weight matrix \(V_j\) is guaranteed to be stable in respect to a sample \(x\) predicted with a confidence of \(\gamma_f(x) > L^{d}\ (\|W_j\|_\infty + \|U_j\|_\infty)\ \|x\|_\infty\prod_{i=1, i\neq j}^{d}\|W_i\|_\infty\).
\end{lemma}

Statements of the kind made above are always theoretical in nature and gauge worst case effects of alterations of a model on the considered outcome. The weak, yet tight, bound on the norm of additive perturbations obtained from masking weights propagates through subsequent statements and thus limit the obtainable guarantees. As the results in the following section of the paper show, however, applying the proposed method yields highly promising results in application.\\
Furthermore, the bounds obtained from above should be considered useful in two regards. Firstly, they yield an easily quantifiable estimate of what can still be guaranteed when applying the proposed method of introducing semi-structured sparsity to a model via maskings. In fact, commonly used regularization techniques such as weight decay or norms on weights in loss functions directly yield smaller bounds on the norms of the mask-induced perturbations. Secondly, understanding in what ways sparse masks affect model performance can guide a practitioner to develop heuristics that minimize the downside effect on (guaranteed) model performance, e.g., by specifically masking weights of low magnitude and by bounding the maximum magnitude of weights to be masked.

\subsection{Semi-Structured Sparse Convolutions}\label{sub:reg_sparse_conv}
\begin{figure}[ht]
	\centering
	\def\svgwidth{0.95\linewidth}
	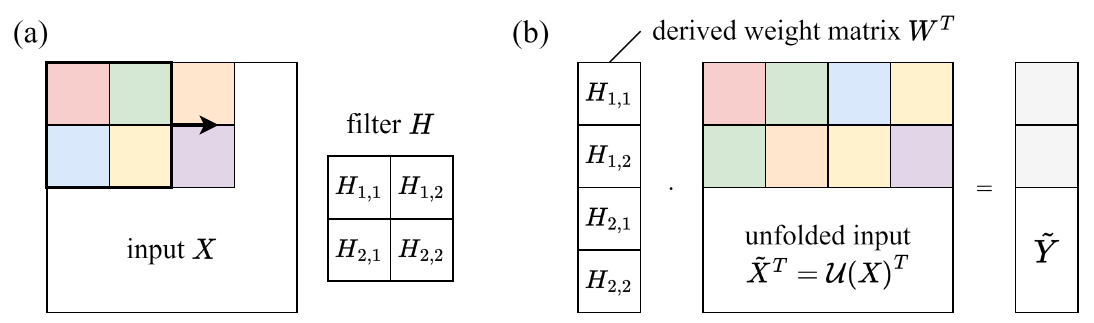
	\caption{Simplified visualizations of convolutions on single channel input \(X\) of unspecified width and height and a single filter \(H\) as (a) ``standard'' convolution with a moving filter and (b) as a matrix product between an unfolded input \(\tilde X\) and a weight matrix \(W\) derived from the filter}
	\label{fig:lin_matrix}
\end{figure}

Since \gls{cv} models commonly rely on two-dimensional convolutions to process the (image) inputs, the application of semi-structured sparsity on convolutions is illustrated in two dimensions. The method extends to other dimensions in an analogue fashion. A discrete two-dimensional convolution with a kernel \(H \in \mathbb R^{c_\text{in}\times h \times w}\) convolves an input tensor \(X \in \mathbb R^{c_{\text{in}}\times b \times d}\) into an output tensor \(y \in \mathbb R^{b \times d}\) assuming zero padding. In the below formulation the functions \(f\) and \(g\) handle the padding for invalid combinations of input values, i.e., out of range values, else return the sum of the two input values:
\begin{equation}\label{equ:std_conv}
	y_{ij} = \sum_{c=1}^{c_\text{in}}\sum_{u=1}^{w}\sum_{s=1}^{h}H_{cus}X_{cf(i,u)g(j, s)}
\end{equation}
Usually such a convolution is conducted with \(c_\text{out}\) kernels to obtain an output \(Y \in \mathbb R^{c_\text{out} \times b \times d}\).
Alternatively, this convolution can also be expressed as a matrix multiplication between the same input \(X \in \mathbb R^{c_{\text{in}\times b \times d}}\) and a weight matrix \(W \in \mathbb R^{c_\text{out}\times (c_\text{in}wh)}\) constructed from the \(c_\text{out}\) kernels in the convolutional layer:
\begin{equation}\label{equ:reform_conv}
	\tilde{Y} = W\mathcal{U}(X)
\end{equation}
The unfold operator \(\mathcal{U}(\cdot)\) turns the matrix \(X\) into a flattened matrix \(\tilde{X} \in \mathbb R^{c_\text{in}wh \times L}\) where \(L = (b + 2p_1 - w - 1)(d + 2p_2 - h - 1)\) denotes the number of blocks in the input. In the case of zero padding, i.e., full padding, the padding sizes in the respective dimensions for uneven kernel dimensions are \(p_1 = \lfloor\frac{w}{2}\rfloor\) and \(p_2 = \lfloor\frac{h}{2}\rfloor\) and thus \(L = bd\). Reshaping \(\tilde{Y}\) recovers the exact same \(Y\) as in (\ref{equ:std_conv}). Note, that the described reformulation of (\ref{equ:std_conv}) as (\ref{equ:reform_conv}) does not change the mathematical operations conducted but rather makes the non-contiguous memory access of the convolution explicit. The cost of the convolution (neglecting the details of data loading) therefore does not change. 
To achieve a 2:4 sparse convolution compatible with the accelerated matrix multiplication for 2:4 sparse matrices a mask \(M \in \{0,1\}^{c_\text{out}\times c_\text{in}wh}\) whose (block) entries are sampled according to a \gls{gs} distribution as described in (\ref{equ:sample_m}) is multiplied entry-wise to the weight matrix.
\begin{equation}\label{equ:mask_conv}
	\tilde{Y} = (M\odot W)\ \mathcal{U}(X)
\end{equation}
The corresponding masking layer therefore learns as many \gls{gs} distributions as there are blocks of four elements in the matrix. Note, that this assumes that the product \(c_\text{out}\cdot c_\text{in}wh\) is a multiple of four, since \(M\) can only contain a multiple of four entries to account for the size of the sparsity pattern. If this is not the case the matrix needs to be augmented column- or row-wise to contain a multiple of 4 entries. A schematic illustration of the two views on convolutions is illustrated in Figure \ref{fig:lin_matrix}. 

\section{Results}\label{sec:results}
In the following the architectures considered for experimental evaluation of the proposed method and the empirical results are reported. The figures were obtained by evaluating the models on the \textbf{ImageNet-1K} multi-class classification challenge \cite{russakovskyImageNetLargeScale2015a}. The details of the dataset and comparisons between reported metrics on validation and test set performance are deferred to Appendix \ref{app:val_performance}. The details of the training procedure employed to train the masking layers of the modified architectures are reported in Appendix \ref{app:training}.

\paragraph{Architectures} To empirically evaluate whether the proposed performance gain via semi-structured sparsity can in fact be achieved without any significant loss in inference performance the following architectures were considered: \textbf{ResNet} \cite{heDeepResidualLearning2016a}, and \textbf{ConvNeXt} \cite{liuConvNet2020s2022}. The models were obtained from PyTorch's Torchvision module \cite{anselPyTorchFasterMachine2024}. In each variant of the models considered, the convolutional layers, unless grouped, were reformulated as in (\ref{equ:reform_conv}) yielding a matrix product in which the weights of \(k\) contiguous blocks of entries could be masked to yield a 2:4 sparse matrix as described in (\ref{equ:mask_conv}). To obtain the mask a trainable masking layer learning \(k\) \gls{gs} distributions modeling the masking choices for the \(k\) blocks per layer was added from which the mask can be drawn. 
Since grouped convolutions constitute sparse operations themselves, they induce highly sparse large weight matrices in a high-level reformulation as described in (\ref{equ:reform_conv}) and would have slowed down model training and inference beyond feasibility on the available hardware while only promising negligible efficiency gains at the same time. They were thus not altered. Likewise, linear dense layers were not considered for a modification of the above kind as they do not contribute to the compute cost of the models significantly. E.g., in ResNet-50, linear layers account for only 0.3\% of total \glspl{flop} \cite{galeStateSparsityDeep2019} despite accounting for roughly \(8\%\) of the parameters of the model \cite{heDeepResidualLearning2016a}. This further shows, that the number of parameters itself is not a reliable measure of the computational cost incurred by a layer.
The pretrained weights of the original architectures were copied into the modified models initializing the choice weights of the masking layers randomly using Glorot initialization \cite{glorotUnderstandingDifficultyTraining2010} and a normal distribution with \(\sigma = 10^{-6}\) for weights and biases respectively. Only the weights associated with the masking layers were configured to be trainable leaving the set of pretrained weights unchanged.

\paragraph{Inference Performance} Table \ref{tab:trained_performance} summarizes the results obtained for the ResNet and ConvNeXt architectures indicating the training effort needed to converge to a top-\(1\) accuracy comparable to or better than the originally reported figures. The results show that both the ResNet-based as well as the ConvNeXt-based architectures converged to the non-sparse performance levels with negligible to no loss in performance, both in top-1 and top-5 accuracy. Convergence was reached for all architectures after training periods of a fraction of the length of the original training periods showing empirically that only a comparatively small share of additional resources is needed to learn the proposed efficiency modification. This neglects the fact that modern training recipes \cite{liuConvNet2020s2022} make heavy use of augmentation diminishing the additionally spent share even further in comparison. Further training beyond convergence to the reported performance, reported in Table \ref{tab:tenth_performance}, showed further improvement both in the top-1 and top-5 accuracy commonly beyond the reported performance for the unmodified networks.

\begin{table}[htbp]
	\caption{Validation classification performance of the 2:4 sparse networks measured as the top-\(k\) accuracy on ImageNet-1K \cite{russakovskyImageNetLargeScale2015a} and the number of epochs needed to converge to a comparable or better top-1 accuracy than reported in contrast to the original number of training epochs}
	\begin{center}
		\begin{tabularx}{\columnwidth}{Ycccccc}
			\toprule
			\multirow{2}{*}[-.25em]{\textbf{Architecture}} & \multicolumn{3}{c}{\textbf{reported}} & \multicolumn{3}{c}{\textbf{2:4 sparse}} \\
			\cmidrule(lr){2-4}\cmidrule(lr){5-7}
			& \textbf{top-1} & \textbf{top-5} & \textbf{epochs} & \textbf{top-1} & \textbf{top-5} & \textbf{epochs} \\
			\midrule
			ResNet-18 & 69.76 & 89.08 & 90 & 70.01 & 88.17 & 1 \\
			ResNet-34 & 73.31 & 91.42 & 90 & 75.33 & 91.19 & 1 \\
			ResNet-50 & 76.13 & 92.86 & 90 & 78.54 & 92.86 & 1 \\
			\midrule
			ConvNeXt-T & 82.52 & 96.15 & 300 & 82.51 & 94.67 & 10 \\
			ConvNeXt-S & 83.62 & 96.65 & 300 & 83.76 & 95.00 & 9 \\
			\bottomrule
		\end{tabularx}
	\end{center}
	\label{tab:trained_performance}
\end{table}

\begin{table}[htbp]
	\caption{Validation classification performance of the 2:4 sparse networks measured as the top-\(k\) accuracy on ImageNet \cite{russakovskyImageNetLargeScale2015a} after spending 10\% of the resources used to initially train the network measured by the number of epochs without data augmentation}
	\begin{center}
		\begin{tabularx}{\columnwidth}{Ycccccc}
			\toprule
			\multirow{2}{*}[-.25em]{\textbf{Architecture}} & \multicolumn{3}{c}{\textbf{reported}} & \multicolumn{3}{c}{\textbf{2:4 sparse}} \\
			\cmidrule(lr){2-4}\cmidrule(lr){5-7}
			& \textbf{top-1} & \textbf{top-5} & \textbf{epochs} & \textbf{top-1} & \textbf{top-5} & \textbf{epochs} \\
			\midrule
			ResNet-18 & 69.76 & 89.08 & 90 & 70.22 & 88.27 & 9 \\
			ResNet-34 & 73.31 & 91.42 & 90 & 75.45 & 91.25 & 9 \\
			ResNet-50 & 76.13 & 92.86 & 90 & 78.78 & 92.96 & 9 \\
			\midrule
			ConvNeXt-T & 82.52 & 96.15 & 300 & 85.63 & 96.09 & 30 \\
			ConvNeXt-S & 83.62 & 96.65 & 300 & 87.53 & 96.71 & 30 \\
			\bottomrule
		\end{tabularx}
	\end{center}
	\label{tab:tenth_performance}
\end{table}

To compare the results of the proposed method to a state of the art method of computing a 2:4 sparse subnetwork two variants of the heuristic proposed in \cite{poolChannelPermutationsSparsity2021}, the so-called efficacy score to evaluate pruning patterns, available via NVIDIA's Apex library \cite{nvidiacorporationApexPyTorchExtension2018} were used in the same regime. To conform to the idea of not altering the original weights the networks were not retrained\footnote{Performance metrics are reported for select models in \cite{poolChannelPermutationsSparsity2021} indicating no performance loss after retraining the pruned models for an additional 100 epochs.} after pruning as originally proposed in \cite{poolChannelPermutationsSparsity2021} and the results are aggregated in Table \ref{tab:apex_performance}. As such, no significant compute resources needed to be spent. It can be observed that for all variants of ResNet and ConvNeXt the loss in performance is significant even in the better performing variant allowing for channel permutations before selecting the 2:4 sparse subnetwork. The method proposed in this paper always manages to learn a significantly better sparse pattern while spending less than a tenth of the resources.

\begin{table}[htbp]
	\caption{Validation classification performance of the 2:4 sparse networks obtained via the apex library \cite{poolChannelPermutationsSparsity2021} measured as the top-\(k\) accuracy on ImageNet \cite{russakovskyImageNetLargeScale2015a}. The networks are compared in two settings disallowing and allowing permutations of the channels before pruning.}
	\begin{center}
		\begin{tabularx}{.8\columnwidth}{Ycccccc}
			\toprule
			& \multicolumn{2}{c}{\textbf{dense}} & \multicolumn{4}{c}{\textbf{2:4 sparse (Apex)}} \\
			\cmidrule{2-3}\cmidrule{4-5}\cmidrule{6-7}
			\textbf{Architecture} & \multicolumn{2}{c}{\textbf{reported}} & \multicolumn{2}{c}{\textbf{not permuted}} & \multicolumn{2}{c}{\textbf{permuted}} \\
			\cmidrule{2-3}\cmidrule{4-5}\cmidrule{6-7}
			& \textbf{top-1} & \textbf{top-5} & \textbf{top-1} & \textbf{top-5} & \textbf{top-1} & \textbf{top-5}\\
			\midrule
			ResNet-18 & 69.76 & 89.08 & 17.20 & 36.13 & 21.48 & 41.76 \\
			ResNet-34 & 73.31 & 91.42 & 43.75 & 68.27 & 49.27 & 73.71 \\
			ResNet-50 & 76.13 & 92.86 & 30.09 & 52.52 & 48.37 & 72.47 \\
			\midrule
			ConvNeXt-T & 82.52 & 96.15 & 72.61 & 90.85 & 75.90 & 92.70 \\
			ConvNeXt-S & 83.62 & 96.65 & 75.07 & 92.49 & 76.44 & 93.22 \\
			\bottomrule
		\end{tabularx}
	\end{center}
	\label{tab:apex_performance}
\end{table}

\section{Discussion}\label{sec:discussion}
Spending only a fraction of the resources invested to pretrain the network the results show that it is possible to learn semi-structured 2:4 sparsity patterns that can accelerate \gls{cnn} inference while not impeding or even improving classification performance. This shows that extending the native support of 2:4 sparse matrix operations to 2:4 sparse convolutional kernels is a highly promising avenue towards more efficiency that is achievable today.

The results presented were obtained using a simple, generic training procedure. However, recent works indicate the high relevance of the training recipe, which could go as far as being the sole reason why (vision) transformers outperformed \glspl{cnn} in image classification tasks in recent years \cite{liuConvNet2020s2022}. The effect of more sophisticated training procedures including, e.g., data augmentation \cite{cubukRandaugmentPracticalAutomated2020, yunCutmixRegularizationStrategy2019,zhangMixupEmpiricalRisk2017,zhongRandomErasingData2020}, needs to be studied offering potential to accelerate convergence and reaching even higher levels of classification performance.
Furthermore, the proposed method does not yet make use of available heuristics with patterns still being randomly initialized. In the case of, e.g., ConvNeXt, in which more than one epoch was needed to converge, a meaningful initialization cheaply obtainable, e.g., \cite{poolChannelPermutationsSparsity2021}, could serve as an improved starting point and reduce lifetime resource spending even further.
Lastly, while working exceptionally well, the work thus far only explores image classification. However, \gls{cv} models are also frequently employed for object detection and segmentation tasks. Future experiments could be aimed at surveying all \gls{cv} tasks relevant for \gls{cv} models employed as backbones in foundation models.
From a theoretical viewpoint more assumptions, e.g., on the distribution of weights could lead to more constrained yet tighter bounds. While losing some generality, this could lead to results that are even more interesting from a practical viewpoint to guide effective trades off between inference acceleration and model performance.

Beyond the aforementioned proposals for future work several additional avenues come into consideration: Firstly, the proposed architectural change introduces the temperature parameter \(\tau\) of the \gls{gs} distribution to the reformulated convolutional models, but the effects on performance and convergence are yet to be studied. Secondly, the work can be extended to cover more models, both convolutional and non-convolutional, as well as other frequently used, costly layer types. Lastly, more detailed insights into what information the network loses when modified as proposed could prove valuable.

\section{Conclusion}
In this paper, a novel method for accelerating inference in \gls{cv} architectures has been presented. By expressing convolutions as semi-structured sparse matrix operations existing hardware accelerations for semi-structured sparsity can be used to directly translate model sparsity into significant savings in regards to data loading and \glspl{flop} spent during inference. The proposed use of semi-structured sparsity patterns bridges the gap between practical requirements induced by compute hardware and the theoretical desire to not limit the choice of sparse models to not affect model performance negatively.

To obtain the sparsity patterns, a semi-structured sparse masking of the pretrained model's weights is learned from the training data optimizing for the same goal as the unchanged model. The resources spent on learning the maskings constitute a fraction of the resources spent during the original training of the model effectively reducing the resources spent during the model's lifetime. At the same time, despite dropping out half of the weights in each convolutional layer, the performance of the model is not affected negatively, even increasing in many instances as the classification experiments conducted show. From a theoretical perspective, the effects of masking the weights of a classifier are quantified in the paper in the form of guarantees on class predictions under maskings. Combining these results with model changes induced by updates of the pretrained weights guarantees for reusing learned sparsity patterns can be derived.

In conclusion, the proposed method demonstrates that extending the support of readily available acceleration techniques to natively support convolutional kernels is a promising avenue to accelerate convolutional models more than two-fold while retaining the pretrained performance.

\section*{Acknowledgments}
I would like to thank Dr.\ Peter Belcák for the insightful discussions on the presented work, and Prof.\ Dr.\ Roger Wattenhofer and the Distributed Computing (DISCO) Group at ETH Zürich for providing the necessary resources without which this work would not have been possible.

\bibliographystyle{abbrv}
\bibliography{main.bib}

\begin{thebibliography}{10}

\bibitem{anselPyTorchFasterMachine2024}
J.~Ansel, E.~Yang, H.~He, N.~Gimelshein, A.~Jain, M.~Voznesensky, B.~Bao, P.~Bell, D.~Berard, E.~Burovski, G.~Chauhan, A.~Chourdia, W.~Constable, A.~Desmaison, Z.~DeVito, E.~Ellison, W.~Feng, J.~Gong, M.~Gschwind, B.~Hirsh, S.~Huang, K.~Kalambarkar, L.~Kirsch, M.~Lazos, M.~Lezcano, Y.~Liang, J.~Liang, Y.~Lu, C.~K. Luk, B.~Maher, Y.~Pan, C.~Puhrsch, M.~Reso, M.~Saroufim, M.~Y. Siraichi, H.~Suk, S.~Zhang, M.~Suo, P.~Tillet, X.~Zhao, E.~Wang, K.~Zhou, R.~Zou, X.~Wang, A.~Mathews, W.~Wen, G.~Chanan, P.~Wu, and S.~Chintala.
\newblock {{PyTorch}} 2: {{Faster Machine Learning Through Dynamic Python Bytecode Transformation}} and {{Graph Compilation}}.
\newblock In {\em Proceedings of the 29th {{ACM International Conference}} on {{Architectural Support}} for {{Programming Languages}} and {{Operating Systems}}, {{Volume}} 2}, pages 929--947, La Jolla CA USA, Apr. 2024. ACM.

\bibitem{blalockWhatStateNeural2020a}
D.~Blalock, J.~J. Gonzalez~Ortiz, J.~Frankle, and J.~Guttag.
\newblock What is the state of neural network pruning?
\newblock {\em Proceedings of machine learning and systems}, 2:129--146, 2020.

\bibitem{brownLanguageModelsAre2020}
T.~B. Brown, B.~Mann, N.~Ryder, M.~Subbiah, J.~Kaplan, P.~Dhariwal, A.~Neelakantan, P.~Shyam, G.~Sastry, A.~Askell, S.~Agarwal, A.~{Herbert-Voss}, G.~Krueger, T.~Henighan, R.~Child, A.~Ramesh, D.~M. Ziegler, J.~Wu, C.~Winter, C.~Hesse, M.~Chen, E.~Sigler, M.~Litwin, S.~Gray, B.~Chess, J.~Clark, C.~Berner, S.~McCandlish, A.~Radford, I.~Sutskever, and D.~Amodei.
\newblock Language {{Models}} are {{Few-Shot Learners}}.
\newblock {\em arXiv preprint arXiv:2005.14165}, July 2020.

\bibitem{bulucParallelSparseMatrixMatrix2012}
A.~Bulu{\c c} and J.~R. Gilbert.
\newblock Parallel {{Sparse Matrix-Matrix Multiplication}} and {{Indexing}}: {{Implementation}} and {{Experiments}}.
\newblock {\em SIAM Journal on Scientific Computing}, 34(4):C170--C191, Jan. 2012.

\bibitem{chaiDeepLearningComputer2021}
J.~Chai, H.~Zeng, A.~Li, and E.~W. Ngai.
\newblock Deep learning in computer vision: {{A}} critical review of emerging techniques and application scenarios.
\newblock {\em Machine Learning with Applications}, 6:100134, Dec. 2021.

\bibitem{cubukRandaugmentPracticalAutomated2020}
E.~D. Cubuk, B.~Zoph, J.~Shlens, and Q.~V. Le.
\newblock Randaugment: {{Practical}} automated data augmentation with a reduced search space.
\newblock In {\em Proceedings of the {{IEEE}}/{{CVF}} Conference on Computer Vision and Pattern Recognition Workshops}, pages 702--703, 2020.

\bibitem{desislavovComputeEnergyConsumption2023}
R.~Desislavov, F.~{Mart{\'i}nez-Plumed}, and J.~{Hern{\'a}ndez-Orallo}.
\newblock Compute and {{Energy Consumption Trends}} in {{Deep Learning Inference}}.
\newblock {\em Sustainable Computing: Informatics and Systems}, 38:100857, Apr. 2023.

\bibitem{galeStateSparsityDeep2019}
T.~Gale, E.~Elsen, and S.~Hooker.
\newblock The state of sparsity in deep neural networks.
\newblock {\em arXiv preprint arXiv:1902.09574}, 2019.

\bibitem{glorotUnderstandingDifficultyTraining2010}
X.~Glorot and Y.~Bengio.
\newblock Understanding the difficulty of training deep feedforward neural networks.
\newblock In {\em Proceedings of the Thirteenth International Conference on Artificial Intelligence and Statistics}, pages 249--256. {JMLR Workshop and Conference Proceedings}, 2010.

\bibitem{grimaldiAcceleratingDeepNeural2023}
M.~Grimaldi, D.~C. Ganji, I.~Lazarevich, and S.~S. Deeplite.
\newblock Accelerating {{Deep Neural Networks}} via {{Semi-Structured Activation Sparsity}}.
\newblock In {\em 2023 {{IEEE}}/{{CVF International Conference}} on {{Computer Vision Workshops}} ({{ICCVW}})}, pages 1171--1180, Paris, France, Oct. 2023. IEEE.

\bibitem{gumbelStatisticalTheoryExtreme1954}
E.~J. Gumbel.
\newblock {\em Statistical Theory of Extreme Values and Some Practical Applications: A Series of Lectures}, volume~33.
\newblock US Government Printing Office, 1954.

\bibitem{hanLearningBothWeights2015}
S.~Han, J.~Pool, J.~Tran, and W.~Dally.
\newblock Learning both weights and connections for efficient neural network.
\newblock {\em Advances in neural information processing systems}, 28, 2015.

\bibitem{heDeepResidualLearning2016a}
K.~He, X.~Zhang, S.~Ren, and J.~Sun.
\newblock Deep residual learning for image recognition.
\newblock In {\em Proceedings of the {{IEEE}} Conference on Computer Vision and Pattern Recognition}, pages 770--778, 2016.

\bibitem{hoeflerSparsityDeepLearning2021}
T.~Hoefler, D.~Alistarh, T.~{Ben-Nun}, N.~Dryden, and A.~Peste.
\newblock Sparsity in deep learning: {{Pruning}} and growth for efficient inference and training in neural networks.
\newblock {\em Journal of Machine Learning Research}, 22(241):1--124, 2021.

\bibitem{hookerHardwareLottery2021}
S.~Hooker.
\newblock The hardware lottery.
\newblock {\em Communications of the ACM}, 64(12):58--65, Dec. 2021.

\bibitem{huLoRALowRankAdaptation2021}
E.~J. Hu, Y.~Shen, P.~Wallis, Z.~{Allen-Zhu}, Y.~Li, S.~Wang, L.~Wang, and W.~Chen.
\newblock {{LoRA}}: {{Low-Rank Adaptation}} of {{Large Language Models}}.
\newblock {\em arXiv preprint arXiv:2106.09685}, Oct. 2021.

\bibitem{huangPruningLargeLanguage2024}
W.~Huang, Y.~Hu, G.~Jian, J.~Zhu, and J.~Chen.
\newblock Pruning {{Large Language Models}} with {{Semi-Structural Adaptive Sparse Training}}.
\newblock {\em arXiv preprint arXiv:2407.20584}, Aug. 2024.

\bibitem{hubaraAcceleratedSparseNeural2021}
I.~Hubara, B.~Chmiel, M.~Island, R.~Banner, J.~Naor, and D.~Soudry.
\newblock Accelerated sparse neural training: {{A}} provable and efficient method to find n: M transposable masks.
\newblock {\em Advances in neural information processing systems}, 34:21099--21111, 2021.

\bibitem{iofinovaHowWellSparse2022}
E.~Iofinova, A.~Peste, M.~Kurtz, and D.~Alistarh.
\newblock How {{Well Do Sparse ImageNet Models Transfer}}?
\newblock In {\em 2022 {{IEEE}}/{{CVF Conference}} on {{Computer Vision}} and {{Pattern Recognition}} ({{CVPR}})}, pages 12256--12266, New Orleans, LA, USA, June 2022. IEEE.

\bibitem{jangCategoricalReparameterizationGumbelSoftmax2017}
E.~Jang, S.~Gu, and B.~Poole.
\newblock Categorical {{Reparameterization}} with {{Gumbel-Softmax}}.
\newblock {\em arXiv preprint arXiv:1611.01144}, Aug. 2017.

\bibitem{justusPredictingComputationalCost2018}
D.~Justus, J.~Brennan, S.~Bonner, and A.~S. McGough.
\newblock Predicting the {{Computational Cost}} of {{Deep Learning Models}}.
\newblock In {\em 2018 {{IEEE International Conference}} on {{Big Data}} ({{Big Data}})}, pages 3873--3882, Seattle, WA, USA, Dec. 2018. IEEE.

\bibitem{liEvaluatingEnergyEfficiency2016}
D.~Li, X.~Chen, M.~Becchi, and Z.~Zong.
\newblock Evaluating the {{Energy Efficiency}} of {{Deep Convolutional Neural Networks}} on {{CPUs}} and {{GPUs}}.
\newblock In {\em 2016 {{IEEE International Conferences}} on {{Big Data}} and {{Cloud Computing}} ({{BDCloud}}), {{Social Computing}} and {{Networking}} ({{SocialCom}}), {{Sustainable Computing}} and {{Communications}} ({{SustainCom}}) ({{BDCloud-SocialCom-SustainCom}})}, pages 477--484, Atlanta, GA, USA, Oct. 2016. IEEE.

\bibitem{liPruningFiltersEfficient2017}
H.~Li, A.~Kadav, I.~Durdanovic, H.~Samet, and H.~P. Graf.
\newblock Pruning {{Filters}} for {{Efficient ConvNets}}.
\newblock {\em arXiv preprint arXiv:1608.08710}, Mar. 2017.

\bibitem{linDynamicModelPruning2020}
T.~Lin, S.~U. Stich, L.~Barba, D.~Dmitriev, and M.~Jaggi.
\newblock Dynamic {{Model Pruning}} with {{Feedback}}.
\newblock {\em arXiv preprint arXiv:2006.07253}, June 2020.

\bibitem{liuConvNet2020s2022}
Z.~Liu, H.~Mao, C.-Y. Wu, C.~Feichtenhofer, T.~Darrell, and S.~Xie.
\newblock A {{ConvNet}} for the 2020s.
\newblock {\em arXiv preprint arXiv:2201.03545}, Mar. 2022.

\bibitem{liuRethinkingValueNetwork2019}
Z.~Liu, M.~Sun, T.~Zhou, G.~Huang, and T.~Darrell.
\newblock Rethinking the {{Value}} of {{Network Pruning}}.
\newblock {\em arXiv preprint arXiv:1810.05270}, Mar. 2019.

\bibitem{loshchilovDecoupledWeightDecay2019}
I.~Loshchilov and F.~Hutter.
\newblock Decoupled {{Weight Decay Regularization}}.
\newblock {\em arXiv preprint arXiv:1711.05101}, Jan. 2019.

\bibitem{maoExploringGranularitySparsity2017}
H.~Mao, S.~Han, J.~Pool, W.~Li, X.~Liu, Y.~Wang, and W.~J. Dally.
\newblock Exploring the {{Granularity}} of {{Sparsity}} in {{Convolutional Neural Networks}}.
\newblock In {\em 2017 {{IEEE Conference}} on {{Computer Vision}} and {{Pattern Recognition Workshops}} ({{CVPRW}})}, pages 1927--1934, Honolulu, HI, USA, July 2017. IEEE.

\bibitem{maoExploringRegularitySparse2017}
H.~Mao, S.~Han, J.~Pool, W.~Li, X.~Liu, Y.~Wang, and W.~J. Dally.
\newblock Exploring the {{Regularity}} of {{Sparse Structure}} in {{Convolutional Neural Networks}}.
\newblock {\em arXiv preprint arXiv:1705.08922}, June 2017.

\bibitem{menghaniEfficientDeepLearning2023}
G.~Menghani.
\newblock Efficient {{Deep Learning}}: {{A Survey}} on {{Making Deep Learning Models Smaller}}, {{Faster}}, and {{Better}}.
\newblock {\em ACM Computing Surveys}, 55(12):1--37, Dec. 2023.

\bibitem{molchanovPruningConvolutionalNeural2017}
P.~Molchanov, S.~Tyree, T.~Karras, T.~Aila, and J.~Kautz.
\newblock Pruning {{Convolutional Neural Networks}} for {{Resource Efficient Inference}}.
\newblock {\em arXiv preprint arXiv:1611.06440}, June 2017.

\bibitem{novaGradientfreeStructuredPruning2023}
A.~Nova, H.~Dai, and D.~Schuurmans.
\newblock Gradient-free structured pruning with unlabeled data.
\newblock In {\em International {{Conference}} on {{Machine Learning}}}, pages 26326--26341. PMLR, 2023.

\bibitem{nvidiacorporationApexPyTorchExtension2018}
{NVIDIA Corporation}.
\newblock Apex ({{A PyTorch Extension}}) --- {{Apex}} 0.1.0 documentation.
\newblock https://nvidia.github.io/apex/, 2018.

\bibitem{nvidiacorporationDALINvidiadalicuda1102024}
{NVIDIA Corporation}.
\newblock {{DALI}} (nvidia-dali-{{cudaX}}.{{X}}), 2024.

\bibitem{parkDeepLearningInference2018}
J.~Park, M.~Naumov, P.~Basu, S.~Deng, A.~Kalaiah, D.~Khudia, J.~Law, P.~Malani, A.~Malevich, S.~Nadathur, J.~Pino, M.~Schatz, A.~Sidorov, V.~Sivakumar, A.~Tulloch, X.~Wang, Y.~Wu, H.~Yuen, U.~Diril, D.~Dzhulgakov, K.~Hazelwood, B.~Jia, Y.~Jia, L.~Qiao, V.~Rao, N.~Rotem, S.~Yoo, and M.~Smelyanskiy.
\newblock Deep {{Learning Inference}} in {{Facebook Data Centers}}: {{Characterization}}, {{Performance Optimizations}} and {{Hardware Implications}}.
\newblock {\em arXiv preprint arXiv:1811.09886}, Nov. 2018.

\bibitem{poolAcceleratingInferenceSparsity2021}
J.~Pool, A.~Sawarkar, and J.~Rodge.
\newblock Accelerating {{Inference}} with {{Sparsity Using}} the {{NVIDIA Ampere Architecture}} and {{NVIDIA TensorRT}}.
\newblock https://developer.nvidia.com/blog/accelerating-inference-with-sparsity-using-ampere-and-tensorrt/, July 2021.

\bibitem{poolChannelPermutationsSparsity2021}
J.~Pool and C.~Yu.
\newblock Channel {{Permutations}} for {{N}}:{{M Sparsity}}.
\newblock {\em Advances in neural information processing systems}, 34:13316--13327, 2021.

\bibitem{radfordLearningTransferableVisual2021a}
A.~Radford, J.~W. Kim, C.~Hallacy, A.~Ramesh, G.~Goh, S.~Agarwal, G.~Sastry, A.~Askell, P.~Mishkin, J.~Clark, G.~Krueger, and I.~Sutskever.
\newblock Learning {{Transferable Visual Models From Natural Language Supervision}}.
\newblock {\em arXiv preprint arXiv:2103.00020}, Feb. 2021.

\bibitem{russakovskyImageNetLargeScale2015a}
O.~Russakovsky, J.~Deng, H.~Su, J.~Krause, S.~Satheesh, S.~Ma, Z.~Huang, A.~Karpathy, A.~Khosla, M.~Bernstein, A.~C. Berg, and L.~{Fei-Fei}.
\newblock {{ImageNet Large Scale Visual Recognition Challenge}}.
\newblock {\em International Journal of Computer Vision}, 115(3):211--252, Dec. 2015.

\bibitem{thompsonComputationalLimitsDeep2023a}
N.~Thompson, K.~Greenewald, K.~Lee, and G.~F. Manso.
\newblock The {{Computational Limits}} of {{Deep Learning}}.
\newblock In {\em Ninth {{Computing}} within {{Limits}} 2023}, Virtual, June 2023. LIMITS.

\bibitem{wenLearningStructuredSparsity2016}
W.~Wen, C.~Wu, Y.~Wang, Y.~Chen, and H.~Li.
\newblock Learning structured sparsity in deep neural networks.
\newblock {\em Advances in neural information processing systems}, 29, 2016.

\bibitem{youDrawingEarlyBirdTickets2022}
H.~You, C.~Li, P.~Xu, Y.~Fu, Y.~Wang, X.~Chen, R.~G. Baraniuk, Z.~Wang, and Y.~Lin.
\newblock Drawing {{Early-Bird Tickets}}: {{Towards More Efficient Training}} of {{Deep Networks}}.
\newblock {\em arXiv preprint arXiv:1909.11957}, Feb. 2022.

\bibitem{yunCutmixRegularizationStrategy2019}
S.~Yun, D.~Han, S.~J. Oh, S.~Chun, J.~Choe, and Y.~Yoo.
\newblock Cutmix: {{Regularization}} strategy to train strong classifiers with localizable features.
\newblock In {\em Proceedings of the {{IEEE}}/{{CVF}} International Conference on Computer Vision}, pages 6023--6032, 2019.

\bibitem{yusterFastSparseMatrix2005}
R.~Yuster and U.~Zwick.
\newblock Fast sparse matrix multiplication.
\newblock {\em ACM Transactions on Algorithms}, 1(1):2--13, July 2005.

\bibitem{zhangMixupEmpiricalRisk2017}
H.~Zhang.
\newblock Mixup: {{Beyond}} empirical risk minimization.
\newblock {\em arXiv preprint arXiv:1710.09412}, 2017.

\bibitem{zhangAdvancingModelPruning2022}
Y.~Zhang, Y.~Yao, P.~Ram, P.~Zhao, T.~Chen, M.~Hong, Y.~Wang, and S.~Liu.
\newblock Advancing model pruning via bi-level optimization.
\newblock {\em Advances in Neural Information Processing Systems}, 35:18309--18326, 2022.

\bibitem{zhongRandomErasingData2020}
Z.~Zhong, L.~Zheng, G.~Kang, S.~Li, and Y.~Yang.
\newblock Random erasing data augmentation.
\newblock In {\em Proceedings of the {{AAAI}} Conference on Artificial Intelligence}, volume~34, pages 13001--13008, 2020.

\end{thebibliography}

\newpage

\appendix

\section{Appendix / Supplemental Material}\label{appendix}

\subsection{Proof of Lemma \ref{lemma:classifier}}
\begin{proof}
	The claim follows from the fact that the Lipschitz constant propagates through each layer of the compositional function.
	Consider a single layer \( f_i(x) = \sigma(W_i x + b_i) \). Since \( \sigma \) is \( L \)-Lipschitz it follows that \(f_i\) is \(L_i\)-Lipschitz with \(L_i := L \|W_i\|\) as indicated below where the first inequality uses the Lipschitz-continuity of \(\sigma\) and the last inequality uses the sub-multiplicative property of norms.
	\begin{align}
		\|\sigma(W_i x_1 + b_i) - \sigma(W_i x_2 + b_i)\| &\leq L \|W_i x_1 + b_i - (W_i x_2 + b_i)\|\\
		&= L \|W_i (x_1 - x_2)\|\\
		&\leq L \|W_i\| \|x_1 - x_2\|\\
		&=: L_i \|x_1 - x_2\|
	\end{align}
	Now, consider the composed function \( f_{k} \circ f_{k-1} \circ ... \circ f_1 \); then:
	\begin{align}
		&\|f_{k} (f_{k-1} \circ ... \circ f_1 (x_1)) - f_{k} (f_{k-1} \circ ... \circ f_1 (x_2))\|\\
		&\leq L_k \|f_{k-1} \circ ... \circ f_1 (x_1) - f_{k-1} \circ ... \circ f_1 (x_2)\|\\
		&\leq L_k L_{k-1} ... L_1 \|x_1 - x_2\|
	\end{align}
	Thus, the function \( f_{k} \circ ... \circ f_1 \) is \( L_{k} L_{k-1} ... L_1 \)-Lipschitz. Finally, since the softmax function \( \text{softmax}(z) \) for a vector \( z \) is a normalized exponential function it is known to be \(1\)-Lipschitz. If \( f_{d} \circ ... \circ f_1 \) is \( L' \)-Lipschitz, where \( L' = L_1 L_2 ... L_{d} \), then the overall classifier \( f(x) \) is consequently \(L_f\)-Lipschitz with \(L_f = L' \cdot 1 = L'\).
\end{proof}

\subsection{Proof of Lemma \ref{lemma:addit_perturb}}
\begin{proof}
	The claim follows from the fact that the masking changes the Lipschitz constant of the masked layer which then propagates through each layer of the compositional function. Let the \(i\)-th layer be the masked layer, i.e., \(W_i' = W_i + \Delta W\) and \( f_i'(x) = \sigma(W_i' x + b_i) \). Since \( \sigma \) is \( L \)-Lipschitz it follows that \(f_i'\) is \((L_i + L\|\Delta W\|)\)-Lipschitz with \(L_i := L \|W_i\|\) as indicated below where the first inequality uses the Lipschitz-continuity of \(\sigma\) and the latter two inequalities use the sub-multiplicative property of norms and the triangle inequality respectively.
	\begin{align}
		\|\sigma(W_i' x_1 + b_i) - \sigma(W_i' x_2 + b_i)\| &\leq L \|W_i' x_1 + b_i - (W_i' x_2 + b_i)\|\\
		&= L \|W_i' (x_1 - x_2)\|\\
		&\leq L \|W_i + \Delta W\| \|x_1 - x_2\|\\
		&\leq (L \|W_i\| + \|\Delta W\|) \|x_1 - x_2\|\\
		&=: (L_i + L\|\Delta W\|) =: L'_i
	\end{align}
	Reusing the results from the proof of Lemma \ref{lemma:classifier} on the Lipschitz constant of a compositional function composed with the softmax function it follows: If \( f_{d} \circ ... \circ f_i \circ... \circ f_1 \) is \( L''\)-Lipschitz, where \( L'' = L_1 ... L_i' ... L_{d} = L_1 ... (L_i + L \|W\|) ... L_{d} = L' + L_1 ... \|\Delta W\| L_d = L' + L\|\Delta W \|\prod_{i=1, i\neq j}^{d}\|W_i\|\) (cf. proof of Lemma \ref{lemma:classifier}), then the overall classifier \( f(x) \) is consequently \(L_f\)-Lipschitz with \(L_f = L'' \cdot 1 = L''\).
\end{proof}

\subsection{Proof of Lemma \ref{lemma:norm_of_bitmask}}
\begin{proof}
	Given the matrix \(W\) and the bit mask \(B\) the additive perturbation can be obtained via the following construction
	\begin{equation}
		(\Delta W)_{ij} := (B_{ij} - 1) W_{ij}
	\end{equation}
	yielding
	\begin{equation}
		W_{ij} + (\Delta W)_{ij} = W_{ij} + (B_{ij} - 1) W_{ij} = \begin{cases}
			W_{ij} - W_{ij} & \text{if}\quad B_{ij} = 0\\
			W_{ij} - 0 & \text{if}\quad B_{ij} = 1
		\end{cases}
	\end{equation}
	which is equivalent to the element-wise product of the matrix with the bit mask:
	\begin{equation}
		B_{ij} W_{ij} = \begin{cases}
			0 & \text{if}\quad B_{ij} = 0\\
			W_{ij} & \text{if}\quad B_{ij} = 1
		\end{cases}
	\end{equation}
	The bound on the infinity norm on \(\Delta W\) follows from the definition of the infinity norm \(\|W\|_\infty = \max_{1\leq j\leq n}\sum_{j=1}^{m}|W_{ij}|\) and the increasing property of sums of non-negative values justifying the inequality below:
	\begin{equation}
		\|\Delta W\|_\infty = \max_{1\leq j\leq n}\sum_{j=1}^{m}|(B_{ij} - 1) W_{ij}| \leq \max_{1\leq j\leq n}\sum_{j=1}^{m}|W_{ij}| = \|W\|_\infty
	\end{equation}
\end{proof}

\subsection{Proof of Lemma \ref{lemma:additive_stability}}
\begin{proof}
	Let \(f(x) = (\text{softmax} \circ f_d \circ ... \circ f_j \circ ... \circ f_1)(x)\) be a compositional classifier of depth \(d\) with \(f_i(x) = \sigma(W_ix + b_i)\). Let \(W_j' = W_j + \Delta W\) be the additive perturbation yielding the perturbed layer \(f_j'(x) = \sigma(W'_jx + b_j)\) and classifier \(f'(x) = (\text{softmax} \circ f_d \circ ... \circ f'_j \circ ... \circ f_1)(x)\).
	Let \(x_{j-1}\) be the input to the \(j\)-th layer, i.e., \(x_{j-1} = (f_{j-1} \circ ... \circ f_1)(x) = f_{j-1, 1}(x)\). Then the the following can be observed about the variance in the \(j\)-th layer, where the first inequality uses the Lipschitz-continuity of \(\sigma\) and the last inequalities use the triangle inequality and the sub-multiplicative property of norms respectively.
	\begin{align}
		\|f_j(x_{j-1}) - f_j'(x_{j-1})\| &= \|\sigma(W_j x_{j-1} + b_j) - \sigma((W_j + \Delta W) x_{j-1} + b_j)\|\\
		&\leq L \|W_j x_{j-1} + b_j - ((W_j + \Delta W) x_{j-1} + b_j)\|\\
		&= L \|W_j x_{j-1} + b_j - (W_j x_{j-1} + b_j) -\Delta W x_{j-1}\|\\
		&\leq L (\|W_j x_{j-1} + b_j - (W_j x_{j-1} + b_j)\| + \|-\Delta W x_{j-1}\|)\\
		&\leq L \|\Delta W\| \|x_{j-1}\|
	\end{align}
	This can be extended to a statement on the entire classifier as follows:
	\begin{align}
		\|f(x) - f'(x)\| &= \|(f_{d, j+1} \circ f_j \circ f_{j-1, 1})(x) - (f_{d, j+1} \circ f_j' \circ f_{j-1, 1})(x)\|\\
		&\leq L^{d-(j+1)+1}\left[\prod_{i=j+1}^{d}\|W_i\|\right] \|f_j(x_{j-1}) - f_j'(x_{j-1})\|\\
		&\leq L^{d-j}\left[\prod_{i=j+1}^{d}\|W_i\|\right]\|\Delta W\| \|x_{j-1}\|\\
		&\leq L^{d-j+1}\left[\prod_{i=j+1}^{d}\|W_i\|\right]\|\Delta W\|\ L^{j-1}\left[\prod_{i=1}^{j-1}\|W_i\|\right]\|x\|\\
		&\leq L^{d}\ \|\Delta W\|\ \|x\|\prod_{i=1, i\neq j}^{d}\|W_i\|
	\end{align}
\end{proof}

\subsection{Proof of Lemma \ref{lemma:combined_effect}}
\begin{proof}
	The setting of the lemma is illustrated below: the classifier \(f(x)\) for which a masking \(B\) has been learned is updated yielding the question what bounds can be obtained for the updated classifier \(f_U(x)\) if masked with the same mask \(B\). 
	\[
	\begin{tikzcd}[column sep=5em]
		f(x) \arrow[r, "U_j"] \arrow[d, "{\Delta W=\mu(B,W_j)}"'] & f_U(x) \arrow[d, "{\Delta V=\mu(B,V_j)}"] \\
		f'(x) \arrow[r, "{U_j+\mu(B, U_j)}"'] & f'_U(x)
	\end{tikzcd}
	\]
	Updating the classifier and then applying the mask yields the following:
	\begin{align}
		(V_j)_{ik} + (\mu(B, V_j))_{ik} &= (W_j)_{ik} + (U_j)_{ik} + (\mu(B, W_j + U_j))_{ik}\\
		&= \begin{cases}
			(W_j)_{ik} + (U_j)_{ik} - (W_j)_{ik} - (U_j)_{ik} & \text{if}\quad B_{ik} = 0\\
			(W_j)_{ik} + (U_j)_{ik} - 0 & \text{if}\quad B_{ik} = 1
		\end{cases}
	\end{align}
	This can alternately be expressed as masking the weights of the classifier first and applying a sparse update \(U_j + \mu(B, U_j)\) instead corresponding to the lower path in the above diagram. The two orders of updates are equivalent:
	\begin{align}
		&(W_j)_{ik} + (\mu(B, W_j))_{ik} + ((U_j)_{ik} + (\mu(B, U_j))_{ik}) \\
		&=\begin{cases}
			(W_j)_{ik} - (W_j)_{ik} + (U_j)_{ik} - (U_j)_{ik} & \text{if}\quad B_{ik} = 0\\
			(W_j)_{ik} - 0 + (U_j)_{ik} - 0 & \text{if}\quad B_{ik} = 1
		\end{cases}
	\end{align}
	All perturbations shown in the diagram are additive perturbations. From the associative property of matrix addition it follows that for any two or more perturbations applied to a weight matrix the induced change can be expressed via a single perturbation. This yields \(W_j + U_j + \Delta V = W_j + (U_j + \Delta V) = W_j + (U_j + \mu(B, W_j + U_j))\). From Lemma \ref{lemma:additive_stability} it follows directly that \(f'_U\) is stable respect to a sample \(x\) predicted with a confidence of \(\gamma_f(x) > L^{d}\ \|U_j + \mu(B, W_j + U_j)\|_\infty\ \|x\|_\infty\prod_{i=1, i\neq j}^{d}\|W_j\|_\infty\). Using the result from Lemma \ref{lemma:norm_of_bitmask} the following bound can be obtained: 
	\begin{align}
		\|U_j + \mu(B, W_j + U_j)\|_\infty &\leq \|U_j\|_\infty + \|\mu(B, W_j + U_j)\|_\infty\nonumber\\
		&\leq \|U_j\|_\infty + \|W_j + U_j\|_\infty \leq \|W_j\|_\infty + 2\|U_j\|_\infty
	\end{align}
	Consider therefore the second, lower path in the above diagram to obtain \(f'_U\) in which \(f(x)\) is masked first and then updated. Observe that
	\begin{equation}
		[U_j+\mu(B, U_j)]_{ik} = \begin{cases}
			(U_j)_{ik} + 0 & \text{if}\quad B_{ik} = 0\\
			(U_j)_{ik} - (U_i)_ik & \text{if}\quad B_{ik} = 1
		\end{cases}
	\end{equation}
	and thus \(\|U_j + \mu(B, W_j + U_j)\|_\infty \leq \|U_j\|_\infty\) (cf. argument in the proof of Lemma \ref{lemma:norm_of_bitmask}). This yields \(W_j + \Delta W + (U_j + \mu(B, W_j + U_j)) = W_j + (\Delta W + U_j + \mu(B, W_j + U_j))\) implying for the stability of \(f'_U\) is stable respect to a sample \(x\) predicted with a confidence of \(\gamma_f(x) > L^{d}\ \|\Delta W + U_j + \mu(B, W_j + U_j)\|_\infty\ \|x\|_\infty\prod_{i=1, i\neq j}^{d}\|W_i\|_\infty\). The following bound obtained from the auxiliary result above yields 
	\begin{equation}
		\|\Delta W + U_j + \mu(B, W_j + U_j)\|_\infty \leq \|\Delta W\|_\infty + \|U_j + \mu(B, W_j + U_j)\|_\infty \leq \|W_j\|_\infty + \|U_j\|_\infty
	\end{equation}
	which is a strict improvement over the naive bound from above for any non-trivial update \(U_j\). It can be concluded that \(f'_U\) is stable respect to a sample \(x\) predicted with a confidence of 
	\begin{equation}
		\gamma_f(x) > L^{d}\ (\|W_j\|_\infty + \|U_j\|_\infty)\ \|x\|_\infty\prod_{i=1, i\neq j}^{d}\|W_i\|_\infty
	\end{equation}
\end{proof}

\subsection{Classification Task and Reported and Validation Classification Performance}\label{app:val_performance}

The pretrained networks were retrained and evaluated on the multi-class classification challenge provided by the \textbf{ImageNet-1K} dataset \cite{russakovskyImageNetLargeScale2015a}. The data set contains approximately 1.2 million training images, 50.000 validation and 100.000 test images of 1000 object classes. To ensure consistency between the performance on the validation set and the reported accuracies on the test set the unmodified architectures were evaluated showing no significant differences (cf. Table \ref{tab:unchanged_performance}). To assess whether reformulating the layers induced changes due to alteration of the floating point arithmetic, likewise comparisons for the unmodified architectures and the modified architectures without masking were conducted showing no change in predictions and performance as expected and are thus not reported.

The below experimental results can be reproduced with the provided code and were obtained on a NVIDIA GeForce RTX\texttrademark\ 3090 with 24 GB of RAM spending less about 3 to 3:30 minutes per variant of ResNet or ConvNeXt respectively.
\begin{table}[htbp]
	\caption{Reported and validation classification performance of the unmodified architectures measured as the top-\(k\) accuracy on ImageNet \cite{russakovskyImageNetLargeScale2015a}}
	\begin{center}
		\begin{tabularx}{.8\textwidth}{Yccccc}
			\toprule
			\multirow{2}{*}[-.25em]{\textbf{Architecture}} & \multirow{2}{*}[-.25em]{\textbf{Parameters}} & \multicolumn{2}{c}{\textbf{reported}} & \multicolumn{2}{c}{\textbf{validation}} \\
			\cmidrule{3-4} \cmidrule{5-6}
			& & \textbf{top-1} & \textbf{top-5} & \textbf{top-1} & \textbf{top-5} \\
			\midrule
			ResNet-18 & 11.7M & 69.76 & 89.08 & 69.69 & 89.06 \\
			ResNet-34 & 21.8M & 73.31 & 91.42 & 73.24 & 91.42 \\
			ResNet-50 & 25.6M & 76.13 & 92.86 & 75.99 & 92.92 \\
			\midrule
			ConvNeXt-T & 28.6M & 82.52 & 96.15 & 82.18 & 95.96 \\
			ConvNeXt-S & 50.2M & 83.62 & 96.65 & 83.26 & 96.94 \\
			\bottomrule
		\end{tabularx}
	\end{center}
	\label{tab:unchanged_performance}
\end{table}

\subsection{Training Procedure}\label{app:training}
The pretrained modified architectures were trained according to a generic training procedure using the AdamW optimizer \cite{loshchilovDecoupledWeightDecay2019} for optimization with an initial learning rate \(\eta = 1.0\), a momentum of \(\beta = 0.9\) and weight decay with a factor of \(\lambda = 10^{-4}\). Training images were random-cropped to a default resolution of \(224^2\), validation images to \(256^2\). The pixel values were normalized with a mean of \(\mu = (0.485, 0.456, 0.406)\) and a standard deviation of \(\sigma = (0.229, 0.224, 0.225)\). Image loading and processing was \gls{gpu}-accelerated using the DALI library \cite{nvidiacorporationDALINvidiadalicuda1102024}.
The learning rate was adjusted by a step scheduler with no warm-up period adjusting the learning rate every 3 epochs by a factor of \(\gamma = 0.1\). The training did not make use of augmentation of the training data showing each sample in every epoch exactly once. All experiments were run on a compute node equipped with 8 processor cores of 8 GB RAM each and a single NVIDIA GeForce RTX\texttrademark\ 3090 with 24 GB of RAM spending roughly from 1-2 hours per epoch for all (row) entries in Table \ref{tab:trained_performance} and \ref{tab:tenth_performance}.

\end{document}